# A two-stage dual-task learning strategy for early prediction of pathological complete response to neoadjuvant chemotherapy for breast cancer using dynamic contrast-enhanced magnetic resonance images

Bowen Jing, Ph.D.[a,b,c], Jing Wang, Ph.D.[a,b,c],*

[a]Department of Radiation Oncology, University of Texas Southwestern Medical Center, 2280 Inwood Rd, Dallas, TX, USA

[b]Advanced Imaging and Informatics for Radiation Therapy (AIRT) Lab, University of Texas Southwestern Medical Center, 2280 Inwood Rd, Dallas, TX, USA

[c]Medical Artificial Intelligence and Automation (MAIA) Lab, University of Texas Southwestern Medical Center, 2280 Inwood Rd, Dallas, TX, USA

*Author for correspondence: jing.wang@utsouthwestern.edu, phone: 214-648-1795

**Abstract**

**Rationale and Objectives:**

Early prediction of pathological complete response (pCR) can facilitate personalized treatment for breast cancer patients. To improve prediction accuracy at the early time point of neoadjuvant chemotherapy, we proposed a two-stage dual-task learning strategy to train a deep neural network for early prediction of pCR using early-treatment magnetic resonance images.

**Methods:**

We developed and validated the two-stage dual-task learning strategy using the dataset from the national-wide, multi-institutional I-SPY2 clinical trial, which included dynamic contrast-enhanced magnetic resonance images acquired at three time points: pretreatment (T0), after 3 weeks (T1), and after 12 weeks of treatment (T2). First, we trained a convolutional long short-term memory network to predict pCR and extract the latent space image features at T2. At the second stage, we trained a dual-task network to simultaneously predict pCR and the image features at T2 using images from T0 and T1. This allowed us to predict pCR earlier without using images from T2.

**Results:**

The conventional single-stage single-task strategy gave an area under the receiver operating characteristic curve (AUROC) of 0.799 for pCR prediction using all the data at time points T0 and T1. By using the proposed two-stage dual-task learning strategy, the AUROC was improved to 0.820.

**Conclusions:**

The proposed two-stage dual-task learning strategy can improve model performance significantly (p=0.0025) for predicting pCR at the early stage (3$^{rd}$ week) of neoadjuvant chemotherapy. The early prediction model can potentially help physicians to intervene early and develop personalized plans at the early stage of chemotherapy.

# 1. Introduction

Neoadjuvant chemotherapy is often used to treat breast cancer to reduce tumor volume, facilitate surgery, and improve breast conservation rates. The I-SPY1 trial (Investigation of Serial Studies to Predict Your Therapeutic Response with Imaging and Molecular Analysis) was a multi-institutional, nationwide phase 1 trial designed to identify early indicators of treatment response based on molecular and imaging data collected from breast cancer patients undergoing neoadjuvant chemotherapy[1-3]. Following the completion of the phase 1 trial, the I-SPY phase 2 trial (I-SPY2) was designed not only to identify early indicators of treatment response but also to evaluate new therapeutic agents for an expanded cohort of high-risk breast cancer patients[4, 5]. In the I-SPY2 trial, each patient received 12 weeks of treatment with one of the new therapeutic agents and was then treated with 4 cycles of Anthracycline. Dynamic contrast-enhanced magnetic resonance (DCEMR) images were acquired at multiple time points (pretreatment, $3^{rd}$ week, $12^{th}$ week, and after treatment) during the treatment. The pathological complete response was determined at the time of surgery, which was approximately 5 months from the start of treatment.

Previous studies on the I-SPY trials have shown that longitudinal functional MR images can be used to evaluate and predict the outcome of therapy for breast cancer patients. For instance, researchers have found a strong correlation between pathological complete response and functional tumor volume measured via DCEMR[6]. Another study found that changes in functional tumor volume may be useful in predicting recurrence-free survival[3]. In addition to tumor volume, multiple handcrafted features were extracted from DCEMR images acquired before, during and after the treatment to predict pathological complete response for 384 patients in the I-SPY 2 trial[7], achieving an area under the receiver operating characteristic curve (AUROC) of 0.81.

Instead of building prediction models using handcrafted features, deep learning-based approaches have been used to predict the pathological complete response (pCR) for patients in the I-SPY clinical trials. By using data acquired before and after treatment, Duanmu et al. developed a deep learning model which achieved an AUROC of 0.83 for the prediction of pCR for patients in the I-SPY1 trial[8]. Recently, Jing et al. developed a model based on convolutional long short-term memory (LSTM) networks to predict pCR using data of the first 3 time points before the completion of treatment, achieving an AUROC of 0.833 for 624 patients in the I-SPY2 clinical trial[9].

Moreover, researchers have also tried to build models to predict the response at earlier time points as early prediction of treatment response is highly desired clinically[10-13]. A study by Liu et al. shows an AUROC of 0.72 using pretreatment MR images from the I-SPY1 clinical trial[14]. In the study by Jing et al., the AUROC was 0.752 using pretreatment images and clinical data[9]. By incorporating MR images at time point T1 (after 3 weeks of treatment), the AUROC improved to 0.799[9]. However, the performance of these models was still worse than the prediction model using MR images at the later time point (12 weeks of treatment) that achieved an AUROC of 0.833.

The aforementioned studies suggest that the data acquired at the later time point can provide a more accurate prediction of pCR and guide treatment. However, the benefits of a patient response-based adaptive treatment strategy may be outweighed by the extended treatment time and increased toxicity when a patient undergoes multiple rounds of treatment with different therapeutic agents. Therefore, it is still of great interest to predict pCR as accurately as possible at early time points (i.e., pretreatment or 3$^{rd}$ week of treatment). In this study, to improve the performance of early prediction of pCR using data acquired at early time points only, we developed and validated a two-stage dual-task learning strategy. Specifically, we designed a deep neural network-based prediction model that can simultaneously predict pCR and the representation of the mid-treatment (i.e., 12$^{th}$ week) image using early time-point (i.e., pretreatment and 3$^{rd}$ week) images and clinical-subtype data. During model training, DCEMR images of three time points (pretreatment, 3$^{rd}$ week, and 12$^{th}$ week) were used in the two-stage training scheme, where the representation of images at the 12$^{th}$ week was learned at the first stage. Then, the learned representation at the 12$^{th}$ week was predicted using only early time-point images (i.e., pretreatment and 3$^{rd}$ week) at the second stage to aid in pCR prediction. Upon completion of the training, the model can be used to predict pCR at an early time point without using images acquired at the 12$^{th}$ week. The performance of the proposed strategy was evaluated using 5-fold cross-validation on a cohort of 624 patients in the I-SPY2 clinical trial.

## 2. Materials and Methods

### 2.1. Dataset:

The I-SPY2 dataset used in this study is publicly available at the Cancer Imaging Archive [15, 16]. In the study, we used DCEMR images, clinical, and cancer subtype data of 624 anonymized patients. The DCEMR images were acquired at three time points before the completion of chemotherapy: pretreatment (T0), after 3 weeks of treatment (T1), and after 12 weeks of treatment (T2). At each time point, the image consists of three channels: early enhancement map, late enhancement map, and functional tumor volume mask (Fig. 1). Clinical data include the type of therapeutic agents and demographic data such as age, race, and ethnicity. Additionally, cancer subtypes from the previous study were also used[17]. Clinical data and cancer subtypes are listed in Table 1. The inclusion criteria and preprocessing of the images can be found in a previous study[9]. Among 624 patients included in this study, there were 213 patients for whom pCR was achieved after chemotherapy.

## 2.2. Two-stage dual-task learning:

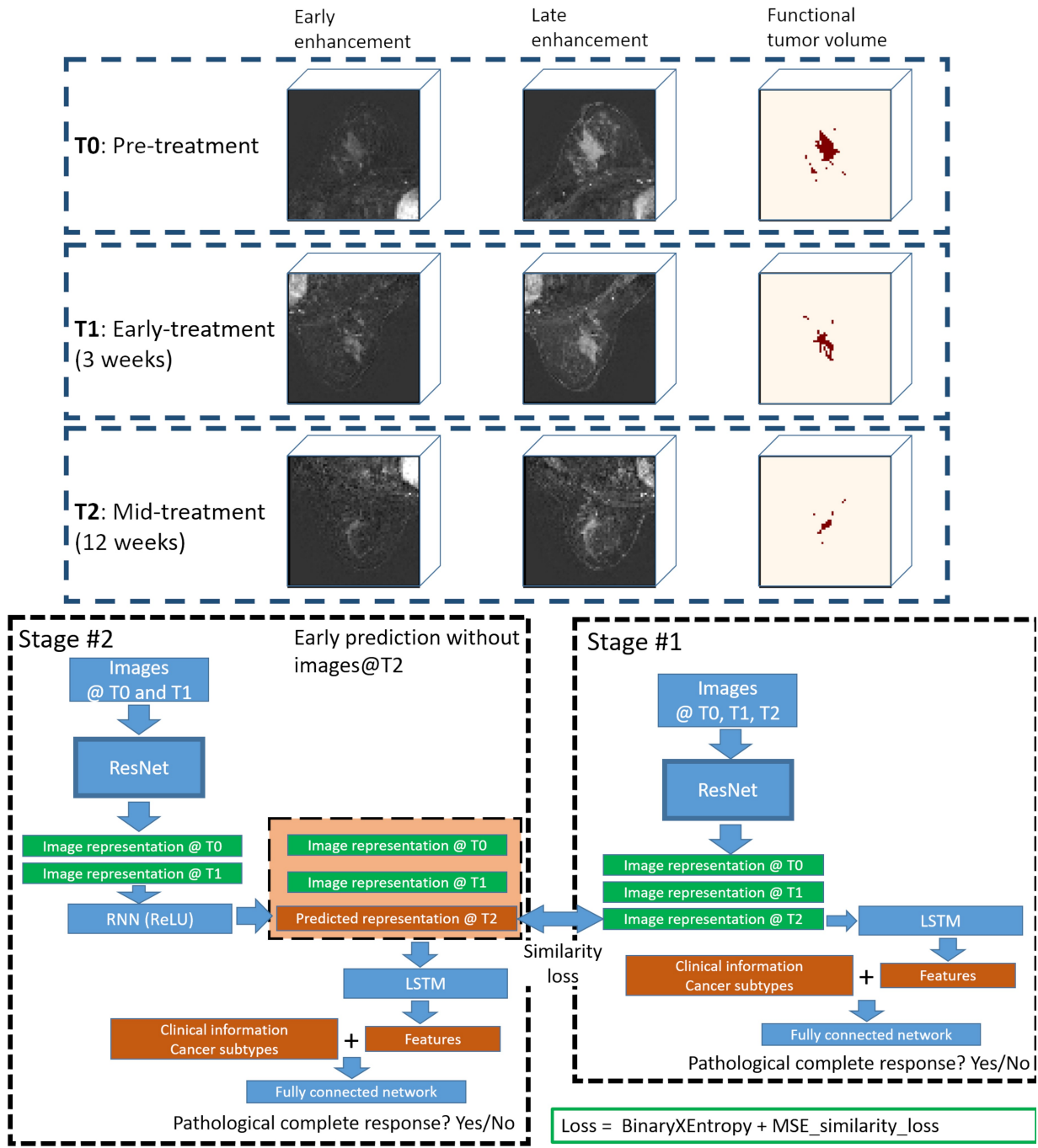

Fig. 1. Two-stage dual-task learning method proposed in this study. The top panel is the dynamic contrast enhanced magnetic resonance image at three time points. ResNet, convolutional residual network. LSTM, long short-term memory network. RNN, recurrent neural network. MSE, mean square error.

TABLE 1. Clinical and cancer subtype data of breast cancer patients

|  |  | Value | Normalization/coding |
|---|---|---|---|
| Clinical | Therapeutic drug | ABT 888, AMG 386, Carboplatin, Ganetespib, Ganitumab, MK-2206, Neratinib, Paclitaxel, Pembrolizumab, Pertuzumab, T-DM1, Trastuzumab | Multi-hot coding (one or several drugs for each patient) |
|  | Age at screening | Minimum: 24 years<br>Maximum: 73 years<br>Mean: ~ 49 years | Z-score normalization |
|  | Race: | American Indian or Alaska Native: 3 patients<br>Asian: 41 patients<br>Black or African American: 73 patients<br>Native Hawaiian or Other Pacific Islander: 3 patients<br>White: 504 patients | One-hot coding |
|  | Ethnicity | Hispanic or Latino: 84 patients<br>Not Hispanic or Latino: 540 patients | One-hot coding |
| Subtype | Hormone receptor (HR) expression | Positive: 338 patients<br>Negative: 286 patients | Positive: 0<br>Negative: 1 |
|  | Human epidermal growth factor receptor-2 (HER2) expression | Positive: 152 patients<br>Negative: 472 patients | Positive: 0<br>Negative: 1 |
|  | MammaPrint (MP) status | Positive: 299 patients<br>Negative: 325 patients | Positive: 0<br>Negative: 1 |
|  | DNA-repair deficiency | Positive: 241 patients<br>Negative: 383 patients | Positive: 0<br>Negative: 1 |
|  | Immune activation status | Positive: 292 patients<br>Negative: 332 patients | Positive: 0<br>Negative: 1 |
|  | BluePrint Luminal index | Basal-type: 315 patients<br>HER2-type: 98 patients<br>Luminal-type: 211 patients | One-hot coding |

Previously, the image at the time point T2 (12 weeks of chemotherapy) was found to be significantly more predictive compared with the imaging features obtained at earlier time points such as T0 and T1[9]. However, 12 weeks into chemotherapy treatment will miss the opportunity for early intervention by switching agents for patients who are not good candidates for early ineffective chemo-agents. In this study, we designed a two-stage dual-task training strategy (Fig. 1) to predict pCR earlier using images at T0 and T1, guided by the image at T2 during the training stage. In the first stage, we trained a convolutional LSTM model to predict pCR and extract the T2 image representation in the latent space. After the completion of the first-stage training, we trained a dual-task convolutional LSTM model to predict pCR and the image representation at T2 simultaneously using only images at T0 and T1. The final model obtained will be used for the early prediction of pCR without using the image at T2. By default, all clinical and cancer subtype data were also used in the training and testing.

### 2.2.1. Stage #1—extraction of image representation at T2

In the first stage of training, the structure of the convolutional LSTM model (Fig. 1) was the same as the one proposed in the previous study[9]. Briefly, a 3-dimensional ResNet18 was used to extract image representations, and the sequence of representations of three time points (T0, T1, and T2) were fed into an LSTM network to get an overall image feature vector. In the end, image features were concatenated with clinical data and cancer subtypes to predict pCR. The binary cross-entropy loss was minimized during training. When the loss reached its minimum on the validation dataset, the model was saved for generating the image representation at T2 in the second stage of training.

### 2.2.2. Stage #2—dual-task training of early prediction network

In the second stage of training, the T2 image representation learned earlier was used as an additional regularization term to guide the prediction. Specifically, a ResNet18 was used to extract image representations from the DCEMR at T0 and T1. Then, a 2-layer recurrent neural network with ReLU activation was used to predict the image representation at T2. The mean square error loss was used to maximize the similarity between the predicted representation and the actual representation generated by the model from stage #1 (Fig. 1). The image representation at T0 and T1, and the predicted representation at T2 were fed into a 2-layer LSTM to get image features similar to those in the first stage. Then, image features were concatenated with clinical data and cancer subtypes to predict pCR. The binary cross-entropy loss and mean square error loss were combined and minimized together during training. The model obtained at this stage can be directly used to predict pCR without using images at T2.

### 2.3. Experimental setup:

The neural network was built using Pytorch 2.0.2 with Python 3.10.8. AdamW[18] was used as the optimizer, with an initial learning rate of 0.001 and a weight decay of 0.01. During model training, a learning rate scheduler was used to reduce the learning rate by a factor of 10 when the validation loss failed to decrease for 20 consecutive epochs. A batch size of 64 was used.

In the first stage of training, the dropout rate of image feature, clinical, and subtype feature vectors was 0.5. Random flip, random rotation in the axial plane, and Gaussian noise (standard deviation = 0.3, mean = 0) were employed to augment image data, as proven effective in the previous study[9]. The rotation angle was randomly sampled from a uniform distribution in the range of -180 to 180 degrees. The flip probability was 0.5.

In the second stage of training, the dropout layers were turned off as the similarity loss provided strong regularization, which prevented overfitting. Random flip and random rotation were used to augment image data during training. It should be noted that only the binary cross-entropy loss was calculated to monitor the performance of the model at the second stage.

**2.4. Training, validation, and testing:**

The model was trained, validated, and tested using a nested 5-fold cross-validation strategy. During the training-validation phase, the three training folds were used to train the neural network, and one fold was used for validation. The best model with the lowest validation loss was saved after each training-validation session. By cycling the training-validation process through all four folds of training data, four models were saved (representing the inner loop of the nested cross-validation). The final pCR prediction for the hold-out testing fold was obtained by averaging the probabilities derived from these four models (i.e., soft voting). It should be noted that only the final model obtained at the second stage was tested. The training, validation, and testing process was repeated for each fold until all five folds had served as the hold-out testing dataset once (representing the outer loop of the nested cross-validation).

Patients' ages were z-score normalized in the training folds to facilitate the training of the deep neural network. The mean and standard deviation of the patients' ages, calculated over the three training folds, were subsequently used to normalize the patients' ages in the validation and testing folds.

The dataset was stratified to ensure that the ratio between pCR-positive and pCR-negative cases remained approximately constant (i.e., 0.5) across all five folds. During training, the weighted random sampler was used to reduce the effect of unbalanced classes (i.e., pCR positive vs.

negative). Each patient in the dataset was weighted by a value of one over the number of patients in the same class. This resulted in the minority class being over-sampled in the training set, which balanced the training set across the two classes.

### 2.5. Comparison study:

Since an earlier prediction of pCR can also be achieved using the conventional single-stage single-task approach, we compared the proposed approach with two conventional methods. For the first conventional method, images at T0 and T1 were used both in training and testing. For the second conventional method, since the LSTM network can take inputs of any length, we trained a convolutional LSTM network using images at three time points (T0, T1, and T2), and predicted pCR without using images at T2 in the testing stage.

### 2.6. Ablation study:

In the default training and testing setup, images at all three time points were used in the first stage, and T0 and T1 images were used in the second stage of training to predict pCR without using images at T2. In addition, we also investigated whether it was feasible to predict pCR using the proposed training strategy with only one of the early time points (T0 or T1). Therefore, we tested the performance of the proposed method when only a single time point was used in the second stage of the proposed two-stage training strategy. Additionally, we also investigated the effect of clinical and subtype data on the performance of the proposed two-stage dual-task learning strategy by excluding one or both data during training and testing.

### 2.7. Evaluation and statistical tests:

The receiver operating characteristic (ROC) curve, area under the ROC curve (AUROC), sensitivity, and specificity were calculated for the entire dataset. Prior to assessing sensitivity and specificity, the probability of a pathological complete response was determined by converting the raw model output score into a probability using a sigmoid function. An operating point of 0.5 probability was used to compute sensitivity and specificity. The AUROCs of models were compared using paired DeLong's test (R programming language, version 4.4.0). The 95% confidence intervals (CIs) for AUROC, sensitivity, and specificity were estimated through 1,000 bootstrap resamples.

3. Results

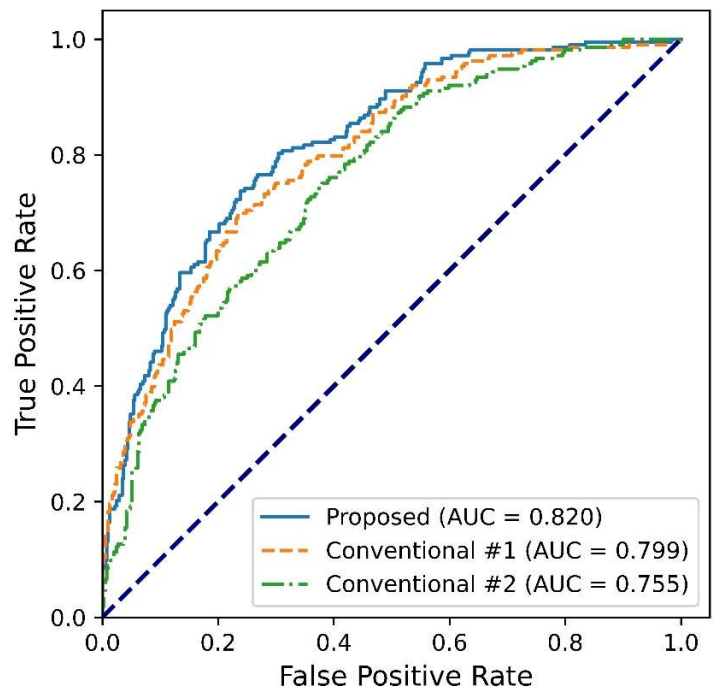

Fig. 2. Receiver operating characteristic curves (ROCs) of the predictions obtained using the proposed two-stage dual-task learning method and the conventional single-stage single-task learning method. Conventional #1 stands for the model that was trained using images at T0 and T1. Conventional #2 stands for the model that was trained using images at T0, T1, and T2, and tested only using images at T0 and T1.

TABLE 2 Prediction performance of models trained using different methods.

| | AUROC [95% CI] | Sensitivity [95% CI] | Specificity [95% CI] |
|---|---|---|---|
| **2-stage dual-task: using T0+T1+clinical+subtypes** | **0.820** [0.787, 0.851] | **0.700** [0.641, 0.760] | **0.779** [0.740, 0.818] |
| 2-stage dual-task without clinical & subtypes | 0.663 [0.618, 0.707] | 0.498 [0.429, 0.563] | 0.762 [0.719, 0.799] |
| 2-stage dual-task without clinical | 0.801 [0.765, 0.836] | 0.676 [0.611, 0.739] | 0.766 [0.726, 0.806] |
| 2-stage dual-task without subtypes | 0.681 [0.641, 0.722] | 0.437 [0.371, 0.500] | 0.800 [0.760, 0.836] |
| 2-stage dual-task: using T0+clinical+subtypes | 0.759 [0.723, 0.795] | 0.545 [0.479, 0.608] | 0.781 [0.742, 0.819] |
| 2-stage dual-task: using T1+clinical+subtypes | 0.807 [0.773, 0.841] | 0.620 [0.559, 0.686] | 0.827 [0.788, 0.862] |
| Conventional: T0+T1+clinical+subtypes | 0.799 [0.765, 0.831] | 0.700 [0.638, 0.764] | 0.754 [0.716, 0.798] |
| Conventional: trained using T0+T1+T2+clinical+subtypes, tested on T0+T1+clinical+subtypes | 0.755 [0.718, 0.792] | 0.254 [0.198, 0.312] | 0.942 [0.918, 0.965] |

For the first conventional method, images at T0 and T1 were used both in training and testing, resulting in an AUROC of 0.799 (Table 2, and Fig. 2). By training the LSTM network with three time points (T0, T1, and T2) and testing it using T0 and T1, the second conventional method gave an AUROC of 0.755 (Table 2, and Fig. 2). Using the proposed two-stage dual-task learning approach, we predicted pCR using images at T0 and T1 only, which gave an AUROC of 0.820 (Table 2, and Fig. 2). DeLong's test indicates that the AUROC of the proposed method is higher than that of the first conventional method, with a p-value of 0.0025, and also significantly higher than the AUROC of the second conventional method (p<0.001).

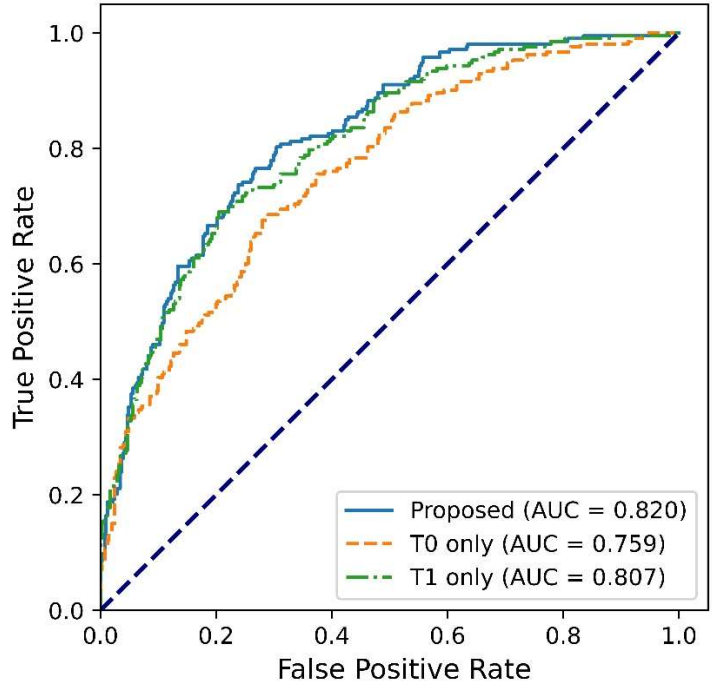

Fig. 3. Receiver operating characteristic curves (ROCs) of the predictions obtained using images at a single time point, T0 or T1. Both models were trained using the proposed two-stage dual-task learning method.

The results of the ablation study are shown in Figure 3 and Table 2. When only images at T0 were used, the AUROC was 0.759, which was significantly lower than the default setup (p < 0.001). When only images at T1 were used, the AUROC was 0.807, which was lower than the default setup (p = 0.0238). By excluding all the clinical and subtype data, we trained and tested the model only using MR images. The AUROC was 0.663 (Table 2 and Fig. 4). When only the clinical data were excluded, the AUROC of the model was 0.801. When we excluded cancer subtype data, the AUROC of the model was 0.681.

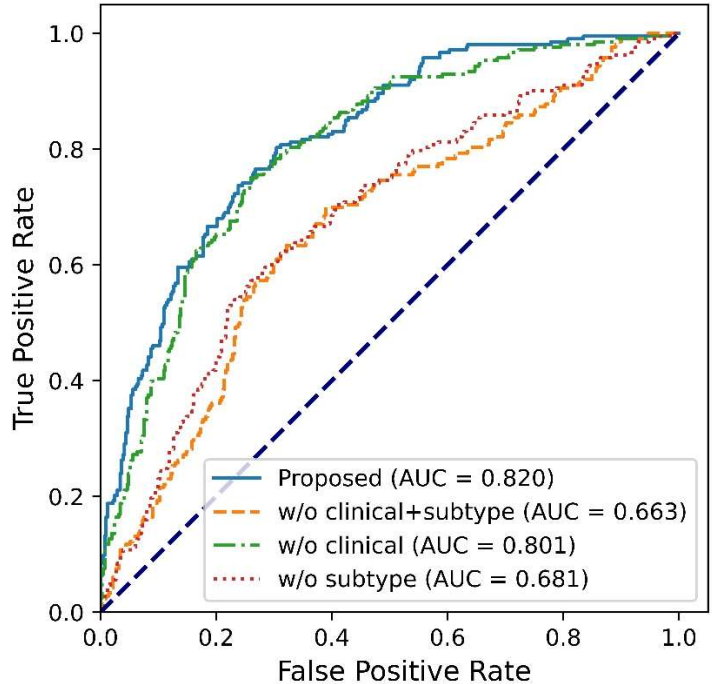

Fig. 4. Receiver operating characteristic curves (ROCs) of the predictions obtained with or without clinical or cancer subtype data. All models were trained using the proposed two-stage dual-task learning method.

## 4. Discussion

The goal of the proposed two-stage dual-task learning strategy is to improve the performance of the model for early prediction of pCR using data acquired at the earlier stage of chemotherapy, i.e., 3 weeks of treatment. While conventional single-stage single-task strategies can provide a pCR prediction with an AUROC of 0.799 by using clinical, subtype data, and images at early time points (T0 and T1), the proposed two-stage dual-task learning strategy has improved the AUROC to 0.820. The improvement over the conventional training strategy is likely attributed to the secondary task where the image representation at T2 was predicted using images at T0 and T1. From the point of view of optimization, the dual-task learning strategy has provided additional regularization, thereby facilitating the optimization process. In fact, the benefits of multi-task learning have been widely demonstrated by published studies on various medical image analysis tasks[19-22]. For treatment response or outcome prediction tasks, tumor segmentation[23, 24], stroma classification[25], and molecular alteration classification[26] have been included as secondary tasks to improve the prediction performance. Compared with these previous studies, our proposed two-stage dual-task learning scheme is unique in terms of the secondary task, which is to predict the image representation at a future time point.

In the default setting of the dual-task learning scheme, we used both the pre- (T0) and early (T1) treatment images to predict the image representation at the $12^{th}$ week (T2). Comparatively, the AUROC dropped significantly if only the image at T0 was used. When only the image at T1 was used, the AUROC only dropped to 0.807, which is not significantly lower than that using images at both T0 and T1 (AUROC = 0.820). These results indicated that the image at the $3^{rd}$ week (T1) was more important for response prediction than the pretreatment image (T0).

As shown in Table 2, the models trained and tested without using cancer subtypes gave AUROCs (<0.7) much lower than those of the model with cancer subtypes (>0.8). These results indicate that cancer subtype data play a significant role in the prediction of treatment response. This is not unexpected, as the therapeutic agents in the I-SPY2 trial were designed to target different molecular pathways, which were categorized into multiple subtypes[17]. Despite this, our study shows that the model that combined cancer subtype data with DCEMR images (AUROC = 0.820) still outperformed the model built solely on clinical and subtype data (AUROC = 0.746[9]). This result indicates that multi-time point DCEMR images can provide indispensable information about the response to chemotherapy.

## 5. Conclusions

In this study, we proposed a two-stage dual-task learning method to build models for the early prediction of response to breast cancer chemotherapy using DCEMR images acquired at early treatment time points (i.e., T0 and T1, pretreatment and 3$^{rd}$ week of treatment). Compared with the conventional single-stage single-task learning approach, the proposed method improved the prediction AUROC from 0.799 to 0.820. This early prediction model can potentially help physicians decide whether to change the treatment plan at an early stage of chemotherapy. Additionally, the proposed two-stage dual-task learning method could potentially be applied to other outcome prediction tasks in which the patient is imaged at multiple time points during the course of treatment.

## 6. Acknowledgement

Funding: This work was supported by the National Institutes of Health [grant number R01CA251792].